\documentclass{article}

\usepackage{arxiv}

\usepackage[utf8]{inputenc}
\usepackage[T1]{fontenc}
\usepackage{libertine}
\usepackage{hyperref}
\usepackage{url}
\usepackage{booktabs}
\usepackage{amsfonts}
\usepackage{nicefrac}
\usepackage{microtype}
\usepackage{graphicx}
\usepackage{doi}
\usepackage{subcaption}
\usepackage{csquotes}
\usepackage{bbm}
\usepackage{dsfont}
\usepackage{algorithm}
\usepackage{algpseudocode}
\usepackage{mathtools}
\usepackage{multirow}
\usepackage{cleveref}

\graphicspath{{graphics/}}

\title{Dynamic Quality-Diversity Search}

\renewcommand{\shorttitle}{\textit{Gallotta, et al.} - Dynamic Quality-Diversity Search}

\author{ 
    \href{https://orcid.org/0000-0001-7578-6173}{\includegraphics[scale=0.06]{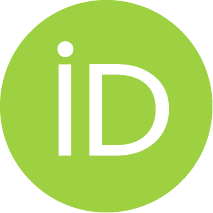}\hspace{1mm}Roberto Gallotta}\\
    	University of Malta\\
    	Msida, Malta \\
    	\texttt{roberto.gallotta@um.edu.mt} \\
	\And
	\href{https://orcid.org/0000-0001-5554-1961}{\includegraphics[scale=0.06]{graphics/orcid.pdf}\hspace{1mm}Antonios Liapis} \\
    	University of Malta\\
    	Msida, Malta \\
    	\texttt{antonios.liapis@um.edu.mt} \\
	\And
	\href{https://orcid.org/0000-0001-7793-1450}{\includegraphics[scale=0.06]{graphics/orcid.pdf}\hspace{1mm}Georgios N. Yannakakis} \\
    	University of Malta\\
    	Msida, Malta \\
    	\texttt{georgios.yannakakis@um.edu.mt} \\
}

\date{}

\hypersetup{
pdftitle={Dynamic Quality-Diversity Search},
pdfsubject={cs.AI},
pdfauthor={Roberto Gallotta, Antonios Liapis, Georgios N. Yannakakis},
pdfkeywords={Quality-Diversity, Dynamic Optimization, Evolutionary Algorithm, Population-based Search},
}

\begin{document}

\maketitle

\begin{abstract}
Evolutionary search via the quality-diversity (QD) paradigm can discover highly performing solutions in different behavioural niches, showing considerable potential in complex real-world scenarios such as evolutionary robotics. Yet most QD methods only tackle static tasks that are fixed over time, which is rarely the case in the real world. Unlike noisy environments, where the fitness of an individual changes slightly at every evaluation, dynamic environments simulate tasks where external factors at unknown and irregular intervals alter the performance of the individual with a severity that is unknown a priori. Literature on optimisation in dynamic environments is extensive, yet such environments have not been explored in the context of QD search. This paper introduces a novel and generalisable Dynamic QD methodology that aims to keep the archive of past solutions updated in the case of environment changes. Secondly, we present a novel characterisation of dynamic environments that can be easily applied to well-known benchmarks, with minor interventions to move them from a static task to a dynamic one. Our Dynamic QD intervention is applied on MAP-Elites and CMA-ME, two powerful QD algorithms, and we test the dynamic variants on different dynamic tasks.
\end{abstract}


\section{Introduction}\label{sec:introduction}
Solving an optimisation problem consists in finding the best performing solution(s) in the domain of interest. While many problems are fixed, many others change in different ways over time. These Dynamic Optimisation Problems (DOPs) \cite{nguyen_continuous_2011, nguyen_dynamic_2009} are of high interest to the research community, as they mirror real-world problems where time affects performance of solutions \cite{cobb_genetic_1993, grefenstette_evolvability_1999}, constraints \cite{cui_memory_2015, nguyen_benchmarking_2009, nguyen_continuous_2012}, or even the bounds of the solution space available to the solving algorithm \cite{abello_adaptive_2011, yaochu_evolutionary_2005}. DOPs are distinct from noisy optimisation problems, where the problem itself is static over time but there is noise in either the evaluation of a solution or in its fitness assignment. While this latter family of optimisation problems also resembles real-world problems, we are more interested in the former group, as time-dependent environments can also be further extended to handle noisy evaluations.

When searching for a solution in a dynamic environment, changes often occur in the fitness landscape. Thus, it is important to be able to detect and adapt to such changes quickly, otherwise previously high-performing solutions may become outdated and hinder the search process \cite{nguyen_continuous_2011}. Applying to DOPs evolutionary algorithms (EAs) \cite{nguyen_evolutionary_2012} and particle swarm optimisation (PSO) \cite{hemlata_dynamic_2012} helps diversify the population, which improves detection of changes in the environment and adaptation. However, to our knowledge, prior applications of such algorithms relied on specific diversification techniques, and did not consider changes in the environment that would alter the measure of diversity itself. On the other hand, quality-diversity (QD) algorithms are a family of EAs that explicitly takes into account the behaviour of solutions to keep a diverse population \cite{pugh_quality_2016}. Additionally, prior work on QD applied to noisy domains \cite{flageat_fast_2020} and with changing featuremaps \cite{fontain_mapping_2019, gallotta_preference_2023} showed promising results. Therefore, we believe QD algorithms provide a solid foundation to solve DOPs, with some critical adaptations.

In this paper, we propose a framework to adapt existing QD algorithms to solve DOPs, improving their performance over their static counterpart. Our Dynamic Quality-Diversity (D-QD), explained in \Cref{sec:dqd}, can search in environments that change at an unknown frequency and with unknown severity, on either the fitness landscape or the behavioural mapping.

The code for this paper is available at \small{\url{https://github.com/gallorob/dynamic-quality-diversity}}.

\section{Related Work}\label{sec:related_work}
This section overviews QD search, with a focus on the two main algorithms our work builds upon, MAP-Elites and CMA-ME, and surveys the field of evolutionary dynamic optimisation.

\subsection{Quality Diversity}\label{subsec:qd}
In evolutionary computation, Quality Diversity (QD) algorithms have been extensively used to efficiently explore the search space of problems by subdividing it into phenotypical \textit{niches} \cite{pugh_quality_2016}. Each solution in a QD problem has an associated fitness, or objective measure, that tells the algorithm the \textit{quality} of the solution, and a set of behavioural characteristics, or \textit{BCs}, based on phenotypical characteristics of the solution. A niche of the search space contains one or more solutions with BCs that fall within a specific interval. By keeping the best solutions (\textit{elites}) based on fitness in niches, the \textit{diversity} of the population is maintained over time.

Maintaining a diverse set of solutions when trying to solve a problem can be extremely valuable, especially in problems where the fitness landscape is deceptive \cite{lehman_abandoning_2011} or where local optima lead solutions away from global optima. QD algorithms are elegant alternatives to algorithms that follow only the fitness signal, and can be applied to both single- and multi-objective optimisation problems \cite{gravina_procedural_2019,pierrot_multi-objective_2022,zhiming_surrogate-asssisted_2019}.

It is equally important for QD search to ensure that the solutions currently present in the archive have the correct properties, both in terms of fitness measure and BCs. QD algorithms may struggle in domains where the fitness has a stochastic component, e.g. when predicted by a surrogate model \cite{gaier_2017_sail}, or where the behaviour of the solution may change unexpectedly, e.g. in real world problems. Such domains can be either \textit{stochastic} \cite{jalili_introduction_2022, neumann_analysis_2019}, where there is a chance that the fitness or the BC value computed is not the true one, or \textit{noisy} \cite{friedrich_compact_2017, gießen_robustness_2016}, where either the fitness value, the BCs, or both are subject to a noisy evaluation \cite{nielsen_adaptive_2019,flageat_fast_2020,wang_adaptive_2022}. In this paper, however, we deal with dynamic environments, in which the true fitness and BCs of a solution change over time due to shifts in the environment during the optimisation process.

\subsubsection{MAP-Elites}\label{subsec:rl_mapelites}
The Multi-dimensional Archive of Phenotypic Elites (MAP-Elites), first introduced in \cite{mouret_illuminating_2015}, is a popular QD algorithm. While many variants of MAP-Elites exist \cite{cully_quality_2018, sfikas_monte_2021}, we modify the simplest version implemented in \cite{mouret_illuminating_2015}. This version of MAP-Elites uses an archive of elites along (usually two or more) dimensions (BCs) and selects parents among these elites uniformly to generate an offspring. The offspring is evaluated and its fitness score and BCs are recorded. If the offspring's BCs would place it in an empty cell, it occupies that cell; otherwise, the offspring is compared against the elite in that cell, and the offspring replaces the occupying elite if the latter's fitness is worse than the offspring's.

\subsubsection{CMA-ME}\label{subsec:rl_cmame}
CMA-ME \cite{fontaine_covariance_2020} is an adaptation of the Covariance Matrix Adaptation Evolution Strategy (CMA-ES) \cite{hansen_completely_2001} to MAP-Elites. As such, it also maintains an archive of elite solutions divided according to their BCs. However, CMA-ME introduces a different sampling strategy via multiple improvement emitters \cite{cully_multi-emitter_2021}. Each improvement emitter works as an instance of CMA-ES, so offspring are generated from a mean solution vector $\mathbf{m}$ and a covariance matrix $\mathbf{C}$.
At the end of a generation, each CMA-ME emitter is updated to explore more promising areas of the search space. While we do not include the complete equations for the sake of brevity, and refer the reader to \cite{hansen_2023_cma} for an in-depth explanation of the CMA-ME algorithm, here we note that two parameters, namely the recombination weights $\mathbf{w}$ and the variance effective selection mass $\mu_{\text{eff}}$ are key parameters to update the mean vector and the covariance matrix.

\subsection{Evolutionary Dynamic Optimisation}\label{subsec:dyn_opt}
Evolutionary Dynamic Optimisation (EDO) is a sub-field of dynamic optimisation that focuses on problems where the objective changes over time, leveraging techniques inspired by biological evolution \cite{nguyen_evolutionary_2012}. Thus, solutions to these problems are not necessarily the best solutions in the whole time window of interest. Instead, to maintain high-performing solutions, an EDO algorithm must be able to both detect and adapt to changes in the environment. 

Changes in the environment may not only affect the fitness landscape, but also the number of variables and constraints. Additionally, such changes can be (a)periodic, occur at a non-fixed rate, and impact with a different severity. Different methods have been introduced to detect such changes, e.g. based on algorithmic behaviour \cite{cobb_investigation_1990, richter_detecting_2009}, which cannot always be applied. More commonly, changes detection is performed via re-evaluation of certain solutions, called \textit{detectors} \cite{abdunnaser_adapting_2006, hu_adaptive_2002, nguyen_continuous_2011, nguyen_solving_2010, zou_evolutionary_2004}, which however introduce additional re-evaluations against the environment \cite{nguyen_evolutionary_2012}. Similarly, to keep high-performing solutions over time, different techniques to enforce diversity in the population have been introduced. Diversity can be either enforced during search \cite{cobb_genetic_1993, morrison_diversity_2004, nguyen_evolutionary_2013}, or introduced when a change in the environment has been detected \cite{moser_simple_2007, richter_learning_2009, richter_memory_2010, riekert_adaptive_2009, vavak_learning_1997}.

While evolutionary algorithms have been used in EDO problems in the past, to our knowledge no QD algorithm has ever been adapted to work in dynamic problems. In this paper, we present a novel way to both detect and to adapt to the shifts in the environment specific to QD algorithms, discussed in Section \ref{sec:dqd}.

\section{Dynamic Quality Diversity}\label{sec:dqd}
Given our goal of keeping the archive of solutions up-to-date when there are (unknown) environment shifts, we identify three core components that can impact the behaviour of any Dynamic QD (D-QD) algorithm: (a) environment shift detection; (b) offspring generation; (c) archive re-evaluation. We include the pseudocode for our D-QD algorithm in \Cref{alg:dqd}.

\begin{algorithm}
\caption{Dynamic QD Search}\label{alg:dqd}
\begin{algorithmic}[1]
    \Require{environment $env$, detection strategy $d$, re-evaluation strategy $r$, algorithm $a$, total timesteps $T_\mathrm{end}$}
    
    \State $A = \emptyset$\Comment{Initialise the archive}
    \While{$t < T_\mathrm{end}$}
        \State $o \leftarrow a(A)$ \Comment{Generate offsprings}
        \State $s_d \leftarrow d(A)$ \Comment{Sample solutions $s_d$ for detection}
        \If{$\mathrm{outdated}(s_d, env, A)$} \Comment{Detect shifts in environment}
            \State $s_r \leftarrow r(A)$ \Comment{Sample solutions $s_r$ for re-evaluation}
            \State  $A \leftarrow A \cup env(s_r)$ \Comment{Update archive with re-evaluated solutions}
        \EndIf
        \State $A \leftarrow A \cup env(o)$ \Comment{Attempt adding offspring to archive}
        \State $a \leftarrow \mathrm{update}(a, A)$ \Comment{Update the algorihtm's emitters}
        \State $env \leftarrow \mathrm{next}(env, t)$ \Comment{Update the environment to the next timestep}
    \EndWhile
    
    \State \Return $A$
\end{algorithmic}
\end{algorithm}

Given the fact that the environment may change at any time, the \emph{environment shift detection} component is vital as it triggers re-evaluation of past solutions (more on this below). We approach the environment shift detection task only from the perspective of its impact on our current archive. If the objective score, BCs, or both, of a solutions have changed from the previously recorded value (stored in the archive), then we assume that there was an environment shift. We do not allow any margin of error in the comparison, therefore we detect a shift regardless of the magnitude of the change. We thus assume our environments are not noisy: alternatively, the comparison between previous and current values would need an additional slack hyperparameter before determining whether a shift has indeed occurred or if the difference was the result of a noisy evaluation of the solution. 

To avoid re-evaluating all individuals in the archive, however, in this paper we test two methods for selecting individuals for detecting environment shifts: (a) oldest solutions or (b) solutions that will be replaced (replacees). For the oldest solutions ($d_O$) method, we assume that solutions that have been in the archive (without having been re-evaluated) are more likely to be outdated. For this method we sample $m$ individuals from the archive according to an exponential function based on their age; thus, solutions evaluated last are less likely to be selected. Selected individuals are re-evaluated and compared against their older values. If any property is different (on any individual), we assume the environment has shifted and perform additional steps in other D-QD components. Regardless of whether the re-evaluated individual has changed values from the previous ones or not, we store the new fitness and BC values and reset the age of the individual. For the replacees ($d_R$) environment detection method, we instead simply use the offspring generated from the offspring generation method (see below) and assess which archive cells they should be assigned to. If these cells are occupied, we take the elites occupying those cells and re-evaluate them. This is beneficial regardless, as ideally an offspring with up-to-date scores should not replace an elite with out-dated scores. Therefore, if any of the re-evaluated replacees' fitness or BCs is different from the same individuals previous values, we detect that the environment has shifted and perform additional steps in other D-QD components. It is worth noting here that (a) re-evaluating replacees based on new offsprings' BC coordinates may lead to fewer samples if many of the cells are unoccupied; (b) re-evaluated replacees may have different BC values and thus move to a different cell which can be occupied, triggering another potential replacement. In this edge case, which we call \emph{cascading re-evaluations}, the occupying elite of the cell that the re-evaluated elite might be entered must be re-evaluated as well---and may also move to a different occupying cell and so on. This means that for $d_R$ the number of re-evaluations may be different from the number of offspring. It is worth noting that, unlike in $d_O$, there is no need to maintain which elites were recently evaluated since any and all candidate elites to be replaced by new offspring will be re-evaluated.

The process of selecting parents and generating offspring is central to all evolutionary and genetic algorithms. Each family of algorithms approaches this problem differently, and thus devising a universal approach for tackling dynamic problems is impossible. In this paper, we adapt two QD algorithms towards D-QD: MAP-Elites, which is a steady-state algorithm where parents are selected among elites, and CMA-ME, which does not have explicit parents but generates offspring from a mean solution vector and a covariance matrix. For both algorithms, we suggest an algorithm-specific offspring generation alternative that we describe under Sections \ref{subsec:dmapelites} and \ref{subsec:dcmame}.

Finally, the issue of re-evaluating the archive of solutions is critical to both the efficiency and the performance of the algorithm. The best-case scenario of re-evaluating all elites in the archive when an environment shift is detected leads to the most up-to-date solutions on average at the cost of compute time. Instead, in this paper we test D-QD algorithms performance when only the replacees are re-evaluated when an environment shift is detected ($e_R$). As discussed above, when offsprings are generated and their BCs computed, they may replace existing elites whose properties may be outdated. In this paper, if an environment shift is detected the elites that may be replaced are re-evaluated and if their values changed they may be moved in the featuremap triggering additional re-evaluations (should they move to occupied cells). We note that, when using $d_R$ the replacees are re-evaluated once (both for environment shift detection and for keeping the archive up-to-date). If instead $d_O$ is used, then (if a shift is detected) replacees will still need to be re-evaluated in this step regardless. Thus, using replacees for detecting shifts may be more computationally efficient than using oldest elites, in the case of frequently shifting environments. However, since the number of re-evaluations with replacees is not bounded (due to cascading re-evaluations) using it for detecting environment shifts is less controllable and may be less reliable.

\subsection{Dynamic MAP-Elites}\label{subsec:dmapelites}
The first QD algorithm we test our approach on is MAP-Elites. As we operate within a dynamic environment implementing the algorithm proposed in \Cref{sec:dqd}, we dub it Dynamic MAP-Elites (D-MAP-Elites).
In the dynamic MAP-Elites variant in this paper, we only modify the parent selection from uniform to biased selection towards parents that have not been selected recently. This is in line with similar selection biases in the literature \cite{cully_quality_2018, sfikas_monte_2021} towards improving coverage of the search space. In our case, we also hypothesise that favouring \enquote{older} elites as parents will create offspring that  may ideally trigger re-evaluations on the replacees.

Parent selection is done via a Beta distribution, controlled by two positive parameters, $\alpha$ and $\beta$, that appear as exponents of the variable and its complement to 1. We change these two values for each cell in the archive at each iteration, driving selection towards certain cells. When a solution in a cell is replaced, we reset its $\alpha$ and $\beta$ values. We initially set $\alpha = \beta = 1$ for all cells in the archive. Then, at the end of each iteration, we update these values by knowing which solutions were picked as parents and which solutions were re-evaluated in case of a detected shift in the environment. We increase $\alpha$ by 1 for all cells, and $\beta$ by 1 in the cells that did not contain a parent solution.

\subsection{Dynamic CMA-ME}\label{subsec:dcmame}
In this paper, we also modify the Covariance Matrix Adaptation MAP-Elites (CMA-ME) to create the variant Dynamic CMA-ME (D-CMA-ME). CMA-ME is an interesting algorithm for adapting to dynamic environments, not only because it has a demonstrated success in QD tasks and beyond \cite{fontaine_covariance_2020, fontaine_illuminating_2021}, but also because it uses a very different offspring generation method that poses interesting problems for D-QD.

In this paper, we modify CMA-ME operating on a hypothesis that generating offspring near older solutions would trigger not only a more up-to-date archive in terms of the offspring but also more relevant re-evaluations of replacess (see above). While CMA-ME generates new offspring towards an area of (potential) improvement, the D-CMA-ME approach in this paper generates new offspring closer to potentially outdated solutions. This is done by modifying the mean vector $\mathbf{m}$ and covariance matrix $\mathbf{C}$. After each iteration, if a shift in the environment has been detected, we first sample $m$ solutions that have not been currently re-evaluated yet uniformly from the archive. We then update the CMA-ME optimiser as described in \cite{hansen_2023_cma} by only using these selected solutions. However, we scale both the weights $\mathbf{w}$ and the variance effective selection mass $\mu_{\text{eff}}$ by a controllable factor $\gamma \leq 1$: $\mathbf{w} \leftarrow \gamma \cdot \mathbf{w}$ and $\mu_{\text{eff}} \leftarrow \gamma \cdot \mu_{\text{eff}}$. The choice of both $m$ and $\gamma$ naturally impacts performance, as they control how much offsprings will be similar to older solutions instead of further exploring the search space. In this work, we set $m$ to the batch size of the emitter, and $\gamma$ to $0.5$.

\section{Experimental Protocol}\label{sec:experiments}
Dynamic QD aims to keep solutions up-to-date with the least re-evaluations possible. There are two extremes to this: either no re-evaluations are ever performed, therefore keeping outdated solutions in the archive, or re-evaluating all solutions whenever an environment shift is detected, which results in the most re-evaluations performed. We treat these two extremes as our baselines. We hypothesise that D-QD can instead keep most of the archive up-to-date with fewer re-evaluations. We formulate the above as more concise research questions:
\begin{enumerate}
    \item[RQ1] How many solutions can Dynamic-QD keep updated?
    \item[RQ2] What are the deviations in objectives and BCs of solutions in the archive compared to the correct values?
    \item[RQ3] What are the trade-offs between re-evaluations and number of up-to-date solutions?
\end{enumerate}

\subsubsection*{Performance Metrics}\label{subsubsec:perf_metrics}
To address the above RQs, it is necessary to compare the output of the algorithm against a perfect-information, perfectly up-to-date system. We leverage this in two ways: (a) \emph{known environment shifts}, i.e. the iteration when the environmental parameters are changed (rather than when an environment shift is detected); (b) an \emph{ideal archive}. The ideal archive is computed on every iteration, and is applied on the current elites of the tested D-QD variant. To compute the ideal archive, all current elites are re-evaluated and assigned to an initially empty equally sized feature map. The resulting ideal archive will have, at most, the same number of elites as in the original algorithm's archive (although the cells they occupy may not be the same). Often, the elites in the ideal archive will be fewer, as re-evaluated BCs place multiple elites on the same cell and thus only one (the fitness according to the up-to-date fitness) survives. We note here that we are, conceptually, only interested in \emph{surviving elites}, i.e. those that would end up in the ideal archive.
With these perfect-information tools, we address RQ1 by measuring the ratio of elites in the actual MAP-Elites feature map that survive in the ideal archive (over all elites). This survival rate ($\%_s$) would ideally be 100\% if all solutions are constantly kept up to date. 

To address RQ2, we observe only \emph{surviving elites} (i.e. those in the ideal archive) and compute the mean squared error (MSE) between their objective score and BC dimensions in the ideal archive and those in the actual MAP-Elites feature map on the same iteration. Averaging across only surviving elites, we compute three metrics ($MSE_{obj}$, $MSE_{BC_1}$, and $MSE_{BC_2}$) as the average values of the respective errors per surviving elite. By averaging only across surviving elites, these measures are not affected by lower or higher survival rates across compared methods, and by keeping each dimension separate we avoid aggregation biases. We also measure the QD score MSE between the actual archive versus the ideal archive on the same iteration ($MSE_{QD}$). The QD score \cite{pugh_quality_2016} is the sum of the objective scores of all elites in the feature map, and is thus aggregating both quality and diversity. A low MSE on all of these four metrics is desirable, as it implies that surviving elites in the ideal archive actually have values close to the correct ones.

Finally, RQ3 addresses the issue of computational efficiency. We would prefer an algorithm that is computationally cheap. However, an algorithm with no re-evaluations would lead to outdated solutions. Thus, the computational cost of keeping solutions up-to-date is more meaningful as a performance metric. Inspired by the Mean Effort Cost of \cite{yannakakis_emerging_2007}, we calculate the number of evaluations needed to reach a survival rate of 50\% and 75\% after each environment shift. We leverage the notion of known environment shifts, and only consider successful the intervals where the survival rate reaches the threshold between known environment shifts. Averaging number of evaluations across all successful intervals, we compute the Mean Evaluation Cost (MEC) and the percentage of successful intervals to triangulate our conclusions. The ideal algorithm would have a low MEC and a large number of successes, i.e. the majority of its stored solutions are up-to-date.

\subsubsection*{Baselines}
To answer the above RQs, we leverage two naïve baselines as the best- and worst-case scenarios for updating the MAP-Elites archive. The \enquote{No Updates} baseline does not detect any environment shift ($d_\emptyset$), thus no re-evaluation is ever performed ($e_\emptyset$). While computationally efficient, it may lead to many outdated solutions that would not survive in the ideal archive. The \enquote{Update All} baseline instead leverages the $d_R$ method and re-evaluates the entire archive when a shift is detected ($e_\forall$). This ensures that most, if not all, solutions are always up-to-date. However, the computational cost of this approach can become prohibitive for large archives, or for frequent environment shifts.

\subsection{Domains}\label{subsec:domains}
In order to assess the performance of D-QD variants and answer the RQs above, we modify established QD benchmarks from the PyRibs library \cite{bryon_pyribs_2023} to make them dynamic. On the one hand, the dynamic sphere environment (\Cref{subsec:sphere}) is simple and for the most part the environment shifts directly impact a solution properties. On the other hand, the dynamic lunar lander environment (\Cref{subsec:lunarlander}) is closer to a real-world environment problem and changes in the environmental parameters may have unforeseen repercussions (either too minor or too catastrophic) on the properties of the solution.

\subsubsection{Dynamic Sphere Environment}\label{subsec:sphere}
\begin{figure}[t]
    \centering
    \includegraphics[width=0.4\columnwidth]{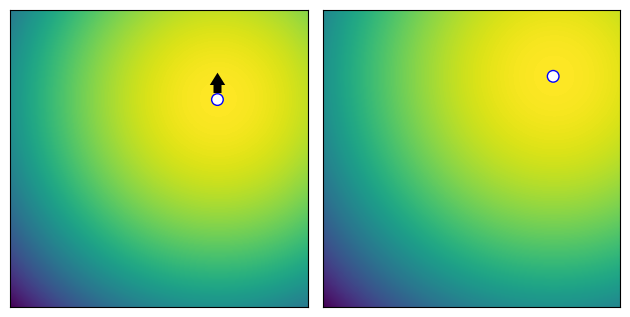}
    \caption{The sphere environment, before a shift (left) and after (right).}\label{img:sphere_domain}
\end{figure}

The sphere environment, first introduced as a QD benchmark in \cite{fontaine_covariance_2020}, is a toy problem for evolutionary optimisation: the goal is to find the global minimum of a sphere function defined in an $n$-dimensional search space. To avoid early convergence in the initial population, the centre of the sphere (i.e. the global minimum) is shifted away from the dimensions' origin (since a typical starting solution for optimisation problems is the zero vector). This shift is usually a fixed offset, however in this paper it is instead time-dependent. 
We modify the existing 100-dimensional sphere environment provided in the PyRibs library to adapt it to our problem requirements. The objective function for the modified sphere environment is $f(\theta)$ in Eq.~\eqref{eq:sphere_objective} where $\theta=\{\theta_1,\ldots,\theta_i,\ldots,\theta_n\}$ is the candidate solution.
\begin{equation}\label{eq:sphere_objective}
    f(\theta) = \sum_{i=1}^{n} (\theta_i - (C_s + \sigma_{obj}))^2,
\end{equation}
\noindent where $n$ is the number of dimensions of the solution ($n = 100$ in this paper), and $\sigma_{obj}$ is the objective shift, initially set to $0$. The centre of the sphere is thus initially set to $C_s=2.048$ based on the original PyRibs implementation. An example of an environment shift is shown, in a simplified 2-D sphere environment, in Fig.~\ref{img:sphere_domain}. 

Behaviour characterisation in the original sphere environment is along two dimensions; in the dynamic version of the environment we simply apply a linear transformation to those BCs as per 
Eq.~\eqref{eq:sphere_bcs}.
\begin{equation}\label{eq:sphere_bcs}
    BC_1(\theta) = \sigma_{BC_1} + \sum_{i=1}^{\lfloor \frac{n}{2} \rfloor} \mathrm{clip}(\theta_i) ;\quad BC_2(\theta) = \sigma_{BC_2} + \sum_{i=\lfloor \frac{n}{2} \rfloor + 1}^{n} \mathrm{clip}(\theta_i),
\end{equation}
\noindent where $\sigma_{BC_1},\sigma_{BC_2}$ are the shift values per BC, and $\mathrm{clip}(\theta_i)$ maintains the solution within the search space bounds.

At each environment shift, the three constants $\sigma_{obj}$, $\sigma_{BC_1}$, $\sigma_{BC_2}$ are modified from their current values (increasing or decreasing). Environment shifts occur at a fixed frequency, after 10 iterations\footnote{Depending on whether MAP-Elites or CMA-ME is used as an underlying algorithm, these 10 iterations may coincide with a different number of generated offspring between shifts.}. In this experiment, every 10 iterations, the previous $\sigma_{obj}$ changes to $\sigma_{obj} \leftarrow \sigma_{obj} + R \cdot U(0, \gamma_{obj})$, where $R$ is a value sampled from a Rademacher distribution (thus set to either $+1$ or $-1$) and $U(0, \gamma_{obj})$ is a value sampled from the uniform distribution between 0 and $\gamma_{obj}$. Similarly, the shift for BCs occurs ever 10 iterations, setting $\{\sigma_{BC_1},\sigma_{BC_2}\} \leftarrow \{\sigma_{BC_1},\sigma_{BC_2}\} + R \cdot \vec{U}(0, \gamma_{BC})$, where $\vec{U}(0, \gamma_{BC})$ is a two-dimensional vector of two values from the uniform distribution between 0 and $\gamma_{BC}$. Thus, while the shifts for both BCs have the same direction (upward or downward), the actual degree of the shift is different between BCs. Based on preliminary tests, the upper limit for the objective and BC values change are $\gamma_{obj}=10$ and $\gamma_{BC}=5$, respectively. Throughout this paper, the initial values of $\sigma_{obj}$, $\sigma_{BC_1}$, $\sigma_{BC_2}$ are zero, and the archive size $100 \times 100$.

The dynamic sphere environment is an appealing environment due to its high level of controllability: shifts are applied directly to the objective and BCs formulas, giving us a perfect benchmark to both validate our code implementation and test our RQs. 

\subsubsection{Dynamic Lunar Lander}\label{subsec:lunarlander}
The lunar lander environment, available in the Gymnasium suite \cite{towers_gymnasium_2023}, is a well-known reinforcement learning physics-based simulation. In this environment, a spaceship (lander) is tasked to safely land on a target location  by controlling only the thrusters on the side of the ship. The lander starts at a random angle and velocity. Each solution is a simple linear policy model, where the action $a$ to be taken based on the current state $\textbf{s}$ of the environment is selected as $a = \mathrm{argmax}(\theta \times \textbf{s})$. We evolve the weights of the $8 \times 4$ decision-making matrix $\theta$.

The objective $f$ of our solution $\theta$ is the cumulative reward given to the lander at each simulation timestep, with an additional reward for crashing ($-100$) or landing successfully ($100$). A solution $\theta$ is considered successful if it has scored a reward of at least $200$. The solution BCs are based on the state of the lander at the timestep when it impacts the ground (regardless whether it is crashing or landing): $BC_1(\theta) \in [-1, 1]$ is the $x$-coordinate of the lander position; and $BC_2(\theta) \in [-3,0]$ is the $y$-coordinate of the lander velocity. Unlike the sphere environment, the objective score and BCs for the solutions are not modified; instead, the conditions of the simulation are modified dynamically and affect the QD components \emph{indirectly}.

\begin{figure}[t]
    \centering
    \includegraphics[width=0.4\columnwidth]{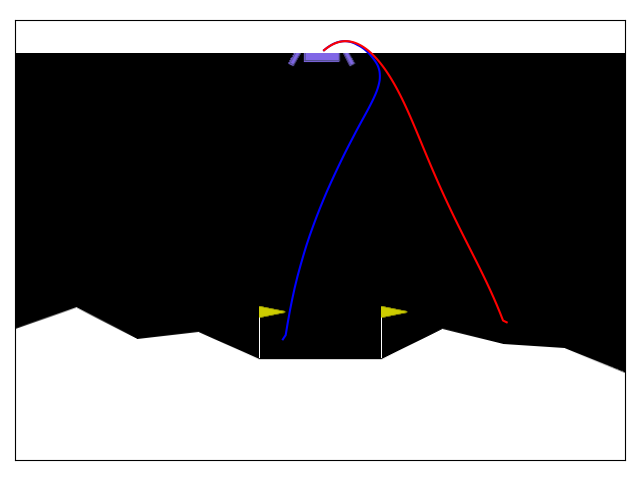}
    \caption{The complete trajectory of the same lander in the lunar lander environment, before a shift (blue) and after (red).}\label{img:lunar_lander_domain}
\end{figure}

Two parameters impact the behaviour of the spaceship throughout the simulation: \textit{wind power} and \textit{wind turbulence}. Fig.~\ref{img:lunar_lander_domain} demonstrates how a shift in wind parameters affects the trajectory taken by the same lander. By changing wind strength and turbulence during the QD search, we obtain a dynamic environment that fits our requirements. The wind power ($\sigma_w$) applies a force $F_w$ to the centre of the lander, whereas the wind turbulence ($\sigma_\tau$) applies a torque $\tau$ to the lander.

The wind and turbulence power are updated when an environment shift occurs. As with the dynamic sphere environment, environment shift occurs at a fixed interval, every 10 iterations. The new values $\sigma_w$, $\sigma_\tau$ are increased or decreased by a small amount and clamped within the recommended value ranges $[0,20]$ and $[0,2]$ respectively. At Every shift, $\sigma_w$ and $\sigma_\tau$ are updated as described in Eq.~\eqref{eq:lunarlander_shifts_update_w} and Eq.~\eqref{eq:lunarlander_shifts_update_T}, respectively.
\begin{align}
    \sigma_{w} &\leftarrow \max( \min( \sigma_{w} + R \cdot U(0, \gamma_w), 20), 0) \label{eq:lunarlander_shifts_update_w}\\
    \sigma_{\tau} &\leftarrow \max( \min( \sigma_{\tau} + R \cdot U(0, \gamma_\tau), 2), 0), \label{eq:lunarlander_shifts_update_T}
\end{align}
\noindent where $R$ is a value sampled from a Rademacher distribution (thus set to either $+1$ or $-1$), $U(0, x)$ is a value sampled from the uniform distribution between 0 and $x$, with $\gamma_w$ and $\gamma_\tau$ being the shift strengths for each parameter. Based on preliminary experiments (summarised below), we set the initial $\sigma_{w}=10$ and $\sigma_\tau=1$ before the first environment shift, and the shift strengths $\gamma_w=0.15$ and $\gamma_\tau=0.1$. QD experiments in the dynamic lunar lander environment use an archive size of $50 \times 50$.

In part, the strength of the dynamic lunar lander environment is that its shifts are indirectly impacting the objective and BCs; at the same time, this means that there is less controllability compared to the dynamic sphere environment. To ensure the effects of the shifts would not result in entirely unpredictable changes in both the objective and BCs, we first analysed the environment itself. We simulated 1000 shifts (following \Cref{eq:lunarlander_shifts_update_w} and \Cref{eq:lunarlander_shifts_update_T}) and measured how different solutions' QD properties change during this simulation. Based on the Pearson correlation and significance testing at $p<0.05$, we find significant correlations between changes in $\sigma_w$ and changes in both objective ($\rho=0.8$) and $BC_1$ ($\rho=-0.9$), while changes in $\sigma_\tau$ are significantly inversely correlated only to $BC_2$ ($\rho=-0.71$). Therefore, we can conclude that there is a linear relationship with all changes in all QD components ($f$, $BC_1$, $BC_2$). To test how localised the change effects were in both objective and BCs, we used the mean Cosine Similarity Entropy \cite{chanwimalueang_cosine_2017} on the absolute changes when an environment shift occurs. We find that environment shifts resulted in non-localised changes in objective ($CSE = 0.04$) and BCs ($CSE = 0.07$ and $CSE = 0.06$ for $BC_1$ and $BC_2$, respectively), i.e.: changes in the environment did affect all solutions indiscriminately.

\section{Results}\label{sec:results}

We report results for each of the RQs defined in Section \ref{sec:experiments}. We focus on the performance of two different offspring generation strategies: \textit{default} uniform selection versus \textit{custom} selection as described in Section \ref{subsec:dmapelites} and Section \ref{subsec:dcmame} for D-MAP-Elites and D-CMA-ME, respectively. We also test the impact of environment shift detection via oldest elites ($d_O$) versus via replacees ($d_R$) as discussed in Section \ref{sec:dqd}. We only re-evaluate replacees ($e_R$) in these experiments, based on preliminary results that indicated a lack of clear advantage of using other strategies. Results are averaged from 10 and 5 independent runs of each algorithm for the dynamic sphere environment and the dynamic lunar lander environment, respectively. Unless otherwise specified in Section \ref{subsec:dmapelites} and Section \ref{subsec:dcmame}, we used the default hyperparameters set by PyRibs \cite{bryon_pyribs_2023}. We report the 95\% confidence interval of these runs in tables. We also compare each variant against both the baselines and the other variants via a Welch's T-Test, applying the Bonferroni correction \cite{dunn_multiple_2012} to $p$-values when performing multiple comparisons. Significance is established at $p<0.05$. We mark with ${}^\dagger$ significant difference between our \enquote{Local Update} variants and the \enquote{No Updates} baseline, and with ${}^\star$ the statistically better results between \enquote{Local Update} variants after Bonferroni correction. Using the performance metrics discussed in Section \ref{subsubsec:perf_metrics}, we attempt to answer each RQ below.

\subsubsection*{RQ1: How many solutions can Dynamic-QD keep updated}
We report the survival rate ($\%_s$) for the baselines and each variant in Table \ref{tab:rq1_all_results}. For the dynamic sphere environment, every D-MAP-Elites and D-CMA-ME variant significantly outperforms their respective \enquote{No Updates} baselines. In D-MAP-Elites, methods that use $d_O$ detection significantly outperform the $d_R$ variants, however the choice of sampling strategy (between default and custom) does not have a statistically significant impact. On the other hand, for D-CMA-ME using the custom sampling for the (generally superior) \{$d_O$, $e_R$\} strategy leads to significant performance gains compared to the default CMA-ME offspring generation method.

For the dynamic lunar lander environment, D-QD variants that use $d_O$ significantly outperform the \enquote{No Updates} baseline, while those that rely on $d_R$ do not. We hypothesise that, while oldest elites are always in place, relying on replacees may not always work for shift detection when most offspring end up in unoccupied cells. On the other hand, for either algorithm there was no statistical impact on performance between custom and default sampling.
\begin{table}[t!]
    \caption{Survival rate ($\%_s$) averaged over all iterations, with 95\% Confidence Interval. The D-QD variants (\enquote{Local Update}) are tested against the \enquote{No Updates} and \enquote{Update all}. Significantly better results between a variant and the \enquote{No Updates} baseline are shown with a ${}^\dagger$, between a variant and the \enquote{Update All} baseline are shown with a ${}^\ddagger$, and between variants are shown with a ${}^\star$.}
    \label{tab:rq1_all_results}
    \centering
    \begin{tabular}{|c|c|c|c|c|}
        \hline
        \multicolumn{5}{|c|}{Dynamic Sphere environment (10 runs)}\\
        \hline
        & Policy & Sample & Strategy& Mean $\%_s$ $\uparrow$\\
        \cline{2-5}
        \multirow{6}{*}{\rotatebox[origin=c]{90}{D-MAP-Elites}}
        & No Updates& default & $d_\emptyset$, $e_\emptyset$& $73 \pm 0.86$ \\
        & Update All& default & $d_R$, $e_\forall$& $99 \pm 0.01$ \\
        \cline{2-5}
        & Local Update & default & $d_{O}$, $e_{R}$  & $87 \pm 0.39^\dagger$ \\
        & Local Update & default & $d_{R}$, $e_{R}$  & $80 \pm 0.44^\dagger$ \\
        & Local Update & custom & $d_{O}$, $e_{R}$  & $87 \pm 0.46^\dagger$ \\
        & Local Update & custom & $d_{R}$, $e_{R}$  & $80 \pm 0.58^\dagger$ \\
        \hline
        \multirow{6}{*}{\rotatebox[origin=c]{90}{D-CMA-ME}}
        & No Updates & default & $d_\emptyset$, $e_\emptyset$& $58 \pm 0.41$ \\
        & Update All & default & $d_R$, $e_\forall$& $100 \pm 0.0$ \\
        \cline{2-5}
        & Local Update & default & $d_{O}$, $e_{R}$  & $90 \pm 0.12^\dagger$ \\
        & Local Update & default & $d_{R}$, $e_{R}$  & $66 \pm 0.76^\dagger$ \\
        & Local Update & custom & $d_{O}$, $e_{R}$  & $94 \pm 0.12^{\dagger,\star}$ \\
        & Local Update & custom & $d_{R}$, $e_{R}$  & $60 \pm 0.71^\dagger$ \\
    \hline \hline
        \multicolumn{5}{|c|}{Dynamic Lunar Lander environment (5 runs)}\\
        \hline
        & Policy & Sample & Strategy& Mean $\%_s$ $\uparrow$\\
        \cline{2-5}
        \multirow{6}{*}{\rotatebox[origin=c]{90}{D-MAP-Elites}}
        & No Updates & default & $d_\emptyset$, $e_\emptyset$ & $59 \pm 6.95$ \\
		& Update All & default & $d_R$, $e_\forall$ & $99 \pm 0.2$ \\
		\cline{2-5}
		& Local Update & default & $d_{O}$, $e_{R}$  & $74 \pm 3.83$ \\
		& Local Update & default & $d_{R}$, $e_{R}$  & $63 \pm 6$ \\
		& Local Update & custom & $d_{O}$, $e_{R}$  & $76 \pm 3.03$ \\
		& Local Update & custom & $d_{R}$, $e_{R}$  & $65 \pm 4.68$ \\
		\hline
		\multirow{6}{*}{\rotatebox[origin=c]{90}{D-CMA-ME}}
        & No Updates & default & $d_\emptyset$, $e_\emptyset$ & $44 \pm 8.52$ \\
		& Update All & default & $d_R$, $e_\forall$ & $100 \pm 0.0$ \\
        \cline{2-5}
		& Local Update & default & $d_{O}$, $e_{R}$  & $78 \pm 0.82^\dagger$ \\
		& Local Update & default & $d_{R}$, $e_{R}$  & $49 \pm 5.36$ \\
		& Local Update & custom & $d_{O}$, $e_{R}$  & $74 \pm 1.66^\dagger$ \\
		& Local Update & custom & $d_{R}$, $e_{R}$  & $51 \pm 4.79$ \\
        \hline
    \end{tabular}
\end{table}

\subsubsection*{RQ2: What are the deviations in objectives and BCs of solutions in the archive compared to the correct values?}
We report the mean MSE values for the baselines and each variant in Table \ref{tab:rq2_all_results}. For the dynamic sphere environment, all variants outperform the \enquote{No Updates} baseline, with the exception of the $\{d_{R}$, $e_{R}\}$ variants in D-MAP-Elites only on the $MSE_{QD}$ metric. For D-MAP-Elites we also find no statistical difference between default and custom sampling. For D-CMA-ME with custom sampling and $\{d_O,e_R\}$ strategy we have MSE gains across the board (i.e. more up-to-date solutions on average) than the respective default sampling.

For the dynamic lunar lander environment, in D-MAP-Elites we find that no variants statistically outperform the \enquote{No Updates} baseline, and again the choice of sampling strategy has no significant impact on performance. For D-CMA-ME, significant improvements are achieved on $MSE_{obj}$ for default sampling and $\{d_O,e_R\}$ over all other D-CMA-ME variants and the \enquote{No Updates} baseline. Overall, it seems that detecting environment shifts with oldest elites is beneficial for minimising deviations from the ideal archive in both D-MAP-Elites and D-CMA-ME.
\begin{table*}[t!]
    \caption{Error for the objective ($MSE_{obj}$), BCs ($MSE_{BC_1}$ and $MSE_{BC_2}$), and QD-score ($MSE_{QD}$) averaged over all iterations, with 95\% Confidence Interval. The D-QD variants (\enquote{Local Update}) are tested against the \enquote{No Updates} and \enquote{Update all}. Significantly better results between a variant and the \enquote{No Updates} baseline are shown with a ${}^\dagger$, between a variant and the \enquote{Update All} baseline are shown with a ${}^\ddagger$, and between variants are shown with a ${}^\star$.}
    \label{tab:rq2_all_results}
    \centering
    \begin{tabular}{|c|c|c|c|c|c|c|c|}
        \hline
        \multicolumn{8}{|c|}{Dynamic Sphere environment (10 runs)}\\ \hline
        & \multicolumn{1}{|c|}{\multirow{2}{*}{Policy}} & \multicolumn{1}{|c|}{\multirow{2}{*}{Sample}} & \multicolumn{1}{|c|}{\multirow{2}{*}{Strategy}} & \multicolumn{4}{|c|}{Mean $\downarrow$} \\ \cline{5-8}
        & \multicolumn{1}{|c|}{} & \multicolumn{1}{|c|}{} & \multicolumn{1}{|c|}{} & \multicolumn{1}{|c|}{$MSE_{obj}$} & \multicolumn{1}{|c|}{$MSE_{BC_1}$} & \multicolumn{1}{|c|}{$MSE_{BC_2}$} & \multicolumn{1}{|c|}{$MSE_{QD}$} \\
        \cline{2-8}
        \multirow{6}{*}{\rotatebox[origin=c]{90}{D-MAP-Elites}}
        & No Updates & default & $d_\emptyset$, $e_\emptyset$ & $6.71 \pm 1.33$ & $2.83 \pm 0.27$ & $26 \pm 2.13$ & $955 \pm 342$ \\
        & Update All & default & $d_R$, $e_\forall$ & $0.076 \pm 0.011$ & $0.048 \pm 0.003$ & $0.35 \pm 0.03$ & $7 \pm 2.5$ \\
        \cline{2-8}
        & Local Update & default & $d_{O}$, $e_{R}$ & $2.87 \pm 0.48^\dagger$ & $1.5 \pm 0.06^\dagger$ & $13.3 \pm 1.3^\dagger$ & $328 \pm 123^\dagger$ \\
        & Local Update & default & $d_{R}$, $e_{R}$ & $3.71 \pm 0.45^\dagger$ & $1.7 \pm 0.07^\dagger$ & $16.8 \pm 1.54^\dagger$ & $471 \pm 164$ \\
        & Local Update & custom & $d_{O}$, $e_{R}$ & $2.89 \pm 0.40^\dagger$ & $1.5 \pm 0.06^\dagger$ & $14 \pm 1.5^\dagger$ & $320 \pm 117^\dagger$ \\
        & Local Update & custom & $d_{R}$, $e_{R}$ & $3.72 \pm 0.8^\dagger$ & $1.8 \pm 0.08^\dagger$ & $17 \pm 1.6^\dagger$ & $481 \pm 172$ \\
        \cline{2-8}
        \hline
        \multirow{6}{*}{\rotatebox[origin=c]{90}{D-CMA-ME}}
        & No Updates & default & $d_\emptyset$, $e_\emptyset$ & $7.2 \pm 0.4$ & $38.9 \pm 1.2$ & $71.6 \pm 1.3$ & $957 \pm 81$ \\
        & Update All & default & $d_R$, $e_\forall$ & $0.0 \pm 0.0$ & $0.0 \pm 0.0$ & $0.0 \pm 0.0$ & $0.0 \pm 0.0$ \\
        \cline{2-8}
        & Local Update & default & $d_{O}$, $e_{R}$ & $4.89 \pm 0.35^\dagger$ & $2.25 \pm 0.03^\dagger$ & $40.5 \pm 2.2^\dagger$ & $3.38 \pm 0.69^\dagger$ \\
        & Local Update & default & $d_{R}$, $e_{R}$ & $6.36 \pm 0.32^\dagger$ & $9.52 \pm 0.23^\dagger$ & $53.35 \pm 1.79^\dagger$ & $56 \pm 31^\dagger$ \\
        & Local Update & custom & $d_{O}$, $e_{R}$ & $1.56 \pm 0.16^{\dagger,\star}$ & $1.43 \pm 0.04 ^{\dagger,\star}$ & $15.1 \pm 1.2 ^{\dagger,\star}$ & $0.87 \pm 0.11 ^{\dagger,\star}$ \\
        & Local Update & custom & $d_{R}$, $e_{R}$ & $3.31 \pm 0.18^\dagger$ & $10.56 \pm 0.53^\dagger$ & $35.7 \pm 1.5^\dagger$ & $12.6 \pm 4.6^\dagger$ \\       
    \hline \hline
        \multicolumn{8}{|c|}{Dynamic Lunar Lander environment (5 runs)}\\ \hline
        & \multicolumn{1}{|c|}{\multirow{2}{*}{Policy}} & \multicolumn{1}{|c|}{\multirow{2}{*}{Sample}} & \multicolumn{1}{|c|}{\multirow{2}{*}{Strategy}} & \multicolumn{4}{|c|}{Mean $\downarrow$} \\ \cline{5-8}
        & \multicolumn{1}{|c|}{} & \multicolumn{1}{|c|}{} & \multicolumn{1}{|c|}{} & \multicolumn{1}{|c|}{$MSE_{obj}$} & \multicolumn{1}{|c|}{$MSE_{BC_1}$} & \multicolumn{1}{|c|}{$MSE_{BC_2}$} & \multicolumn{1}{|c|}{$MSE_{QD}$} \\
        \cline{2-8}
        \multirow{6}{*}{\rotatebox[origin=c]{90}{D-MAP-Elites}}
        & No Updates & default & $d_\emptyset$, $e_\emptyset$ & $310 \pm 97$ & $0.13 \pm 0.04$ & $0.6 \pm 0.04$ & $13728 \pm 5224$ \\
        & Update All & default & $d_R$, $e_\forall$ & $25 \pm 9.2$ & $0.003 \pm 0.001$ & $0.07 \pm 0.02$ & $98 \pm 35$ \\
        \cline{2-8}
        & Local Update & default & $d_{O}$, $e_{R}$ & $228 \pm 69$ & $0.05 \pm 0.01$ & $0.54 \pm 0.04$ & $5107 \pm 2519$ \\
        & Local Update & default & $d_{R}$, $e_{R}$ & $274 \pm 78$ & $0.09 \pm 0.02$ & $0.61 \pm 0.04$ & $10508 \pm 3862$ \\
        & Local Update & custom & $d_{O}$, $e_{R}$ & $218 \pm 47$ & $0.05 \pm 0.01$ & $0.6 \pm 0.04$ & $6133 \pm 2818$ \\
        & Local Update & custom & $d_{R}$, $e_{R}$ & $222 \pm 70$ & $0.07 \pm 0.02$ & $0.62 \pm 0.03$ & $11768 \pm 4651$ \\
        \hline
        \multirow{6}{*}{\rotatebox[origin=c]{90}{D-CMA-ME}}
        & No Updates & default & $d_\emptyset$, $e_\emptyset$ & $203 \pm 44$ & $0.2 \pm 0.06$ & $0.7 \pm 0.06$ & $34120 \pm 21689$ \\
        & Update All & default & $d_R$, $e_\forall$ & $0.08 \pm 0.04$ & $0.0 \pm 0.0$ & $0.0 \pm 0.0$ & $0.8 \pm 1.1$ \\
        \cline{2-8}
        & Local Update & default & $d_{O}$, $e_{R}$ & $83 \pm 7 ^{\dagger,\star}$ & $0.02 \pm 0.004^{\dagger,\star}$ & $0.42 \pm 0.02 ^{\dagger,\star}$ & $1401 \pm 216^\dagger$ \\
        & Local Update & default & $d_{R}$, $e_{R}$ & $182 \pm 18.5$ & $0.09 \pm 0.02$ & $0.66 \pm 0.03$ & $27775 \pm 12487$ \\
        & Local Update & custom & $d_{O}$, $e_{R}$ & $183 \pm 27$ & $0.03 \pm 0.004^\dagger$ & $0.6 \pm 0.06$ & $4801 \pm 1994$ \\
        & Local Update & custom & $d_{R}$, $e_{R}$ & $208 \pm 24$ & $0.09 \pm 0.02$ & $0.72 \pm 0.02$ & $16219 \pm 7547$ \\
        \hline
    \end{tabular}
\end{table*}

\subsubsection*{RQ3: What are the trade-offs between re-evaluations and number of up-to-date solutions?}\label{subsec:rq4}
Table \ref{tab:rq4_all_results} summarises the MEC required to reach a survival rate ($\%_s$) over 50\% and 75\%. Notably, the \enquote{No Updates} baseline struggles to surpass a 50\% survival rate in the CMA-ME variant for both environments, based on the success ratio in parentheses. In contrast, the most computationally expensive \enquote{Update All} baseline consistently achieves the challenging $75\%$ survival threshold, albeit with additional evaluations. For a similar success ratio as \enquote{Update All} in the dynamic sphere environment ($99.8\%$), the best D-MAP-Elites variant (with $\{d_R,e_R\}$ strategy regardless of sampling method) has a fraction of the MEC (12\% that of \enquote{Update All}), while the best D-CMA-ME variant with the same success ratio (with $\{d_O,e_R\}$ strategy and custom sampling) achieves it with 40\% the MEC of \enquote{Update All}. This holds for both survival thresholds in this environment. More efficient algorithms, which use $d_R$ and thus do not need the additional onus of evaluating both replacees and oldest elites, do not manage to reach a survival threshold of 75\% as often (especially in the D-CMA-ME variant). In the more challenging dynamic lunar lander environment, D-QD variants fall short of matching the success of the \enquote{Update All} baseline at a 75\% survival rate threshold. However, alternatives utilising the oldest elites for shifts detection reach between double and nearly 10 times the successes of respective $d_R$ variants for 75\% thresholds. For 50\% survival thresholds, the same success ratio as \enquote{Update All} (99\%) can be reached by the best D-MAP-Elites model (with $\{d_O,e_R\}$ strategy and default sampling) at 30\% the MEC of \enquote{Update All}, and for D-CMA-ME at 68\% the MEC of \enquote{Update All} (with $\{d_O,e_R\}$ strategy and default sampling).

From this analysis we can conclude that using dynamic QD can keep solutions up-to-date at a fraction of the computation effort of re-evaluating all solutions whenever an environment shift is detected. Especially detecting environment shifts via oldest elites may be slightly more computationally heavy, but leads to more successes in keeping the archive updated (with much higher success rates for a survival threshold).
\begin{table}[t!]
    \centering
    \caption{Mean Evaluation Cost (MEC) with a $\%_s$ threshold of $50\%$ and $75\%$, averaged over all iterations, with 95\% Confidence Interval. In parentheses is the percentage of instances where the respective survival threshold was reached. The D-QD variants (\enquote{Local Update}) are tested against the \enquote{No Updates} and \enquote{Update all}. Significantly better results between a variant and the \enquote{No Updates} baseline are shown with a ${}^\dagger$, between a variant and the \enquote{Update All} baseline are shown with a ${}^\ddagger$, and between variants are shown with a ${}^\star$.}
    \label{tab:rq4_all_results}
    \begin{tabular}{|c|c|c|c|r@{\;}l|r@{\;}l|}
        \hline
        \multicolumn{8}{|c|}{Dynamic Sphere environment (10 runs)}\\ \hline
        & \multicolumn{1}{|c|}{\multirow{2}{*}{Policy}} & \multicolumn{1}{|c|}{\multirow{2}{*}{Sample}} & \multicolumn{1}{|c|}{\multirow{2}{*}{Strategy}} & \multicolumn{4}{|c|}{MEC $\downarrow$ (successes $\uparrow$)} \\ \cline{5-8}
        & \multicolumn{1}{|c|}{} & \multicolumn{1}{|c|}{} & \multicolumn{1}{|c|}{} & \multicolumn{2}{|c|}{$\%_s \geq 50\%$} & \multicolumn{2}{|c|}{$\%_s \geq 75\%$} \\
        \cline{2-8}
        \multirow{6}{*}{\rotatebox[origin=c]{90}{D-MAP-Elites}}
        & No Updates & default & $d_\emptyset$, $e_\emptyset$ & $100\pm0.0$ & $(99.8\%)$ & $100\pm0.0$ & $(29.7\%)$ \\
        & Update All & default & $d_R$, $e_\forall$ & $1941\pm5.5$ & $(99.8\%)$ & $1941\pm5.5$ & $(99.8\%)$ \\
        \cline{2-8}
        & Local Update & default & $d_{O}$, $e_{R}$ & $315\pm0.2^\ddagger$ & $(99.8\%^\dagger)$ & $315\pm0.2^\ddagger$ & $(99.7\%^\dagger)$ \\
        & Local Update & default & $d_{R}$, $e_{R}$ & $230\pm0.2^\ddagger$ & $(99.8\%^\dagger)$ & $224\pm0.1^\ddagger$ & $(83.1\%^\dagger)$ \\
        & Local Update & parents & $d_{O}$, $e_{R}$ & $315\pm0.2^\ddagger$ & $(99.8\%^\dagger)$& $315\pm0.2^\ddagger$ & $(99.8\%^\dagger)$\\
        & Local Update & parents & $d_{R}$, $e_{R}$ & $230\pm0.2^\ddagger$ & $(99.8\%^\dagger)$& $222\pm0.1^\ddagger$ & $(81.5\%^\dagger)$\\
        \hline 
        \multirow{6}{*}{\rotatebox[origin=c]{90}{D-CMA-ME}}
        & No Updates & default & $d_\emptyset$, $e_\emptyset$ & $800\pm0.0$ & $(74.6\%)$ & $800\pm0.0$ & $(7.1\%)$ \\
        & Update All & default & $d_R$, $e_\forall$ & $5805\pm6.6$ & $(99.8\%)$ & $5805\pm6.6$ & $(99.8\%)$ \\
        \cline{2-8}
        & Local Update & default & $d_{O}$, $e_{R}$ & $2464\pm0.5$ & $(99.8\%^\dagger)$ & $2464\pm0.5^\ddagger$ & $(99.8\%^\dagger)$ \\
        & Local Update & default & $d_{R}$, $e_{R}$ & $1777\pm0.6^\ddagger$ & $(98.6\%^\dagger)$ & $1631\pm0.5^\ddagger$ & $(19.4\%)$ \\
        & Local Update & custom & $d_{O}$, $e_{R}$ & $2280\pm0.7$ & $(99.8\%^\dagger)$ & $2280\pm0.7^\ddagger$ & $(99.8\%^\dagger)$ \\
        & Local Update & custom & $d_{R}$, $e_{R}$ & $1585\pm0.5^\ddagger$ & $(74.5\%)$& $1533\pm1.6^\ddagger$ & $(16.0\%)$\\
    \hline \hline
        \multicolumn{8}{|c|}{Dynamic Lunar Lander environment (5 runs)}\\ \hline
        & \multicolumn{1}{|c|}{\multirow{2}{*}{Policy}} & \multicolumn{1}{|c|}{\multirow{2}{*}{Sample}} & \multicolumn{1}{|c|}{\multirow{2}{*}{Strategy}} & \multicolumn{4}{|c|}{MEC $\downarrow$ (successes $\uparrow$)} \\ \cline{5-8}
        & \multicolumn{1}{|c|}{} & \multicolumn{1}{|c|}{} & \multicolumn{1}{|c|}{} & \multicolumn{2}{|c|}{$\%_s \geq 50\%$} & \multicolumn{2}{|c|}{$\%_s \geq 75\%$} \\
        \cline{2-8}
        \multirow{6}{*}{\rotatebox[origin=c]{90}{D-MAP-Elites}}
        & No Updates & default & $d_\emptyset$, $e_\emptyset$ & $50\pm0.0$ & $(72.5\%)$ & $50\pm0.0$ & $(22.2\%)$ \\
        & Update All & default & $d_R$, $e_\forall$ & $522\pm18.3$ & $(99.0\%)$ & $522\pm18.3$ & $(99.0\%)$ \\
        \cline{2-8}
        & Local Update & default & $d_{O}$, $e_{R}$ & $160\pm0.9^\ddagger$ & $(99.0\%^\dagger)$ & $142\pm1.6^\ddagger$ & $(46.9\%^\dagger)$ \\
        & Local Update & default & $d_{R}$, $e_{R}$ & $96\pm1.0^\ddagger$ & $(74.1\%)$& $80\pm0.8^\ddagger$ & $(22.0\%)$\\
        & Local Update & custom & $d_{O}$, $e_{R}$ & $163\pm0.9^\ddagger$ & $(98.0\%^\dagger)$ & $144\pm1.1^\ddagger$ & $(52.1\%^\dagger)$ \\
        & Local Update & custom & $d_{R}$, $e_{R}$ & $97\pm0.8^\ddagger$ & $(83.8\%)$ & $82\pm0.4^\ddagger$ & $(26.1\%)$ \\
        \hline 
        \multirow{6}{*}{\rotatebox[origin=c]{90}{D-CMA-ME}}
        & No Updates & default & $d_\emptyset$, $e_\emptyset$ & $400\pm0.0$ & $(40.0\%)$ & $400\pm0.0$ & $(10.3\%)$ \\
        & Update All & default & $d_R$, $e_\forall$ & $1584\pm32.7$ & $(99.0\%)$ & $1584\pm32.7$ & $(99.\%)$ \\
        \cline{2-8}
        & Local Update & default & $d_{O}$, $e_{R}$ & $1077\pm10.3$ & $(99.0\%^\dagger)$ & $1062\pm9.1$ & $(87.6\%^\dagger)$ \\
        & Local Update & default & $d_{R}$, $e_{R}$ & $1030\pm12.3^\ddagger$ & $(48.1\%)$ & $701\pm10.3^\ddagger$ & $(9.1\%)$ \\
        & Local Update & custom & $d_{O}$, $e_{R}$ & $1432\pm15.5$ & $(99.0\%^\dagger)$ & $1292\pm5.5$ & $(48.3\%^\dagger)$ \\
        & Local Update & custom & $d_{R}$, $e_{R}$ & $911\pm9.2^\ddagger$ & $(45.5\%)$& $668\pm5.5^\ddagger$ & $(10.7\%)$\\
        \hline
    \end{tabular}
\end{table}

\section{Discussion}\label{sec:discussions}
From the results presented in Section \ref{sec:results}, we observe that our proposed Dynamic QD variants consistently outperform the \enquote{No Updates} baseline on all metrics\textemdash except on MEC, where \enquote{No Updates} however often fails to reach the desired survival ratio. As expected, the best-case \enquote{Update All} baseline outperforms our Dynamic QD variants at the cost of extensive re-evaluations. 

When looking at the best performing D-QD variants, they are able to increase the survival percentage of \enquote{No Updates} by 14\% and 37\%  on the sphere environment, and by 16\% and 34\% on the lunar lander environment, with D-MAP-Elites and D-CMA-ME, respectively. The trade-off between survival, accuracy, and computation cost is however significant as discussed extensively in Section \ref{subsec:rq4}. Overall, we find that using a different subset of individuals (oldest elites) to detect environment shifts leads to more robust behaviour. However, there is no consistent conclusion that can be drawn regarding the impact of different hyperparameters; especially the custom offspring generation pipelines presented in this paper seem to have less of an impact than expected across the different performance metric analysed in Section \ref{sec:results}.

We also identify a number of limitations of the proposed Dynamic QD framework. Results show that the performance of a D-QD variant is largely domain-dependent, rendering it unclear whether a single D-QD hyperparameter configuration can be considered superior to others in untested environments\textemdash a clear example of no free lunch. Additionally, based on how environment shifts are implemented, there is the possibility that previous environment states can be revisited by the optimiser, effectively setting the correct objective and BCs of outdated solutions in the archive. This impacts performance, especially for the \enquote{No Updates} case, as all metrics we report are a direct consequence of solutions being up-to-date. In EDO literature, such a problem can be framed as periodicity in the environment changes, thus, the exploration of memory strategies \cite{richter_learning_2009} to save older solutions and explicitly re-introduce them in the population when the same environment configuration is met again can be included in future work. Another limitation is that the current D-QD framework does not handle noisy domains at all, which implies that any kind of slightly different objective or BC for a solution would be considered as an environment change. In real-world scenarios, it is entirely possible that measurements are not perfectly accurate or perfectly reproducible. Dynamic environments can therefore also be noisy, even if we make a distinction between the two for the purposes of this paper. Equipping our framework with the capability to handle such cases would be an obvious improvement: a possible approach to this is to leverage the recent deep grid archive \cite{flageat_fast_2020}, where multiple elites are kept in each niche. Finally, a limitation of the current implementation lies in the choice of performance metrics for our experiments: our metrics all rely on the supposed knowledge of an ideal, up-to-date archive, which we can create as we have access to the updated environment. This may not be the case in a real-world problem, where the environment may change mid-evaluation or in uneven intervals, rendering the concept of an ideal archive impossible. 

Overall, this project sets the foundation for more ambitious work where changes in the environment reflect real-world issues, such as deterioration of a machine in evolutionary robotics problems. More interestingly, we are interested in addressing changes in the preferences of a human user in AI-assisted design tools \cite{liapis_mixed-initiative_2016}. Being able to correctly identify when a substantial shift in human preferences occurred would be of highest priority, whereas the re-evaluation can be performed via a surrogate model of the user. Such a system can then be used in design-assistance tasks, recommendation systems, realistic simulators, and supply chain management systems. MAP-Elites has already demonstrated its ability to inspire users in creative tasks, working alongside a human who can define the BCs and objectives of their problem at hand \cite{sfikas_design_2023}. Equipping the MAP-Elites algorithms with the ability to handle shifting designer goals and preferences \cite{gallotta_preference_2023,liapis_designer_2013} which are not explicitly stated to the computer is an ambitious avenue for future work in Dynamic QD.

\section{Conclusion}\label{sec:conclusion}
This paper introduced a novel approach to dynamic optimisation which leverages QD algorithms, and applied it to MAP-Elites and CMA-ME algorithms. The proposed Dynamic QD methodology was tested on two well-known QD benchmarks, the sphere environment and the lunar lander, which were adapted to be dynamic over time. The many experiments with multiple hyperparameter setups and baselines in two environments did not reveal a clear-cut optimal Dynamic QD variant. However, the efficiency of Dynamic QD search on different performance metrics highlights the trade-off between accuracy and computation costs. We expect that this work will inspire research on the optimisation of dynamic environments under the powerful lens of quality-diversity.

\section*{Acknowledgements}
    This project has received funding from the European Union’s Horizon 2020 programme under grant agreement No 951911.

\end{document}